\documentclass{article}

\usepackage[final]{neurips_2023}

\usepackage[utf8]{inputenc} 
\usepackage[T1]{fontenc}    
\usepackage{hyperref}       
\usepackage{url}            
\usepackage{booktabs}       
\usepackage{amsfonts}       
\usepackage{nicefrac}       
\usepackage{microtype}      
\usepackage{xcolor}         
\usepackage{graphicx}
\usepackage{subcaption}
\usepackage{booktabs}
\usepackage{afterpage}
\usepackage{wrapfig}

\usepackage[utf8]{inputenc}
\usepackage[hindi,english]{babel}
\def\bng{\bnwxi}
\font\bnwxi=bangwd10

\title{From Classification to Generation: Insights into Crosslingual Retrieval Augmented ICL}

\author{Xiaoqian Li$^{1,3}$ \qquad Ercong Nie$^{1, 2}$ \qquad Sheng Liang$^{\dag}$$^{1,2}$ \\
$^{1}$Center for Information and Language Processing (CIS), LMU Munich, Germany \\
$^{2}$ Munich Center for Machine Learning (MCML), Germany \\
$^{3}$ Academy of Cyber \\
\texttt{Xiaoqian.Li@campus.lmu.de} \\
\texttt{\{nie, shengliang\}@cis.lmu.de}}

\begin{document}

\maketitle

\begin{abstract}
The remarkable ability of Large Language Models (LLMs) to understand and follow instructions has sometimes been limited by their in-context learning (ICL) performance in low-resource languages. To address this, we introduce a novel approach that leverages cross-lingual retrieval-augmented in-context learning (CREA-ICL). By extracting semantically similar prompts from high-resource languages, we aim to improve the zero-shot performance of multilingual pre-trained language models (MPLMs) across diverse tasks. Though our approach yields steady improvements in classification tasks, it faces challenges in generation tasks. Our evaluation offers insights into the performance dynamics of retrieval-augmented in-context learning across both classification and generation domains.
\end{abstract}

\section{Introduction}
In recent years, the field of Natural Language Processing (NLP) has undergone transformative advances, driven primarily by deep transformer techniques~\citep{Vaswani2017AttentionIA, devlin-etal-2019-bert, Radford2019LanguageMAGPT}. The emergence of Large Language Models (LLMs) such as GPT-3~\citep{Brown2020LanguageMA} and GPT-4~\citep{openai2023gpt4} has increased the ability of these models to understand and execute instructions, marking a significant milestone in the field of in-context learning (ICL). These models exhibit exceptional skills in tasks like text classification and generation. They cater to a wide array of applications across various languages, with the primary beneficiaries being languages like English~\citep{conneau-etal-2020-unsupervised, raffel2020exploring, Radford2019LanguageMAGPT}. Benchmarks like XTREME~\citep{Hu2020XTREMEAM} and BUFFET~\citep{asai2023buffet} further validate their capabilities. However, several low-resource languages, with Bangla as a notable example, face inherent challenges, primarily due to the limited availability of pretraining corpora~\citep{artetxe-schwenk-2019-massively, hangya-etal-2022-improving,sazzed-2020-cross}.

\begin{figure*}[t]
	\centering
	\includegraphics[width=1\linewidth]{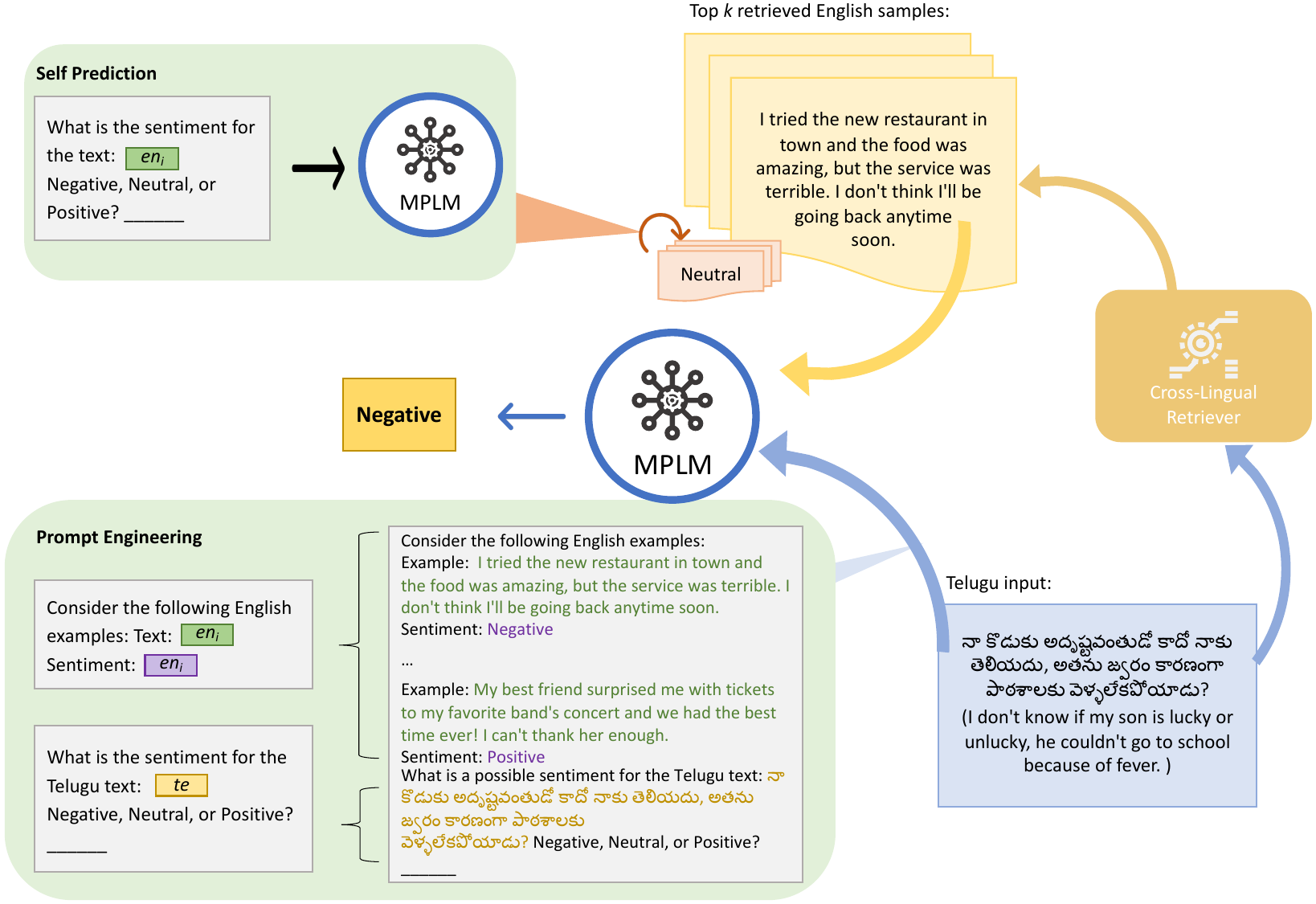}
	\caption{Detailed overview of the CREA-ICL pipeline for LRLs: (a) An LRL input is used as a query for the cross-lingual retriever, which then retrieves the most semantically similar HRL sample from the HRL corpus. The associated label is either taken directly from the corpus (labeled setting) or determined by self-prediction (unlabeled setting). (b) Next, this HRL sample, its label, and the original input are combined to create a retrieval-augmented input for MPLM to make prediction.}
	\label{fig:parc}
\end{figure*}

Despite its vast native speaker base, Bangla's representation in NLP remains constrained. This limitation is attributed to its linguistic complexity, the dearth of labeled datasets, and issues like data duplication~\citep{das2010phrase, das-gamback-2014-identifying}. While traditional machine learning techniques have made progress in Bangla NLP tasks, the potential of harnessing the capabilities of the latest LLMs for ICL remains to be fully tapped~\citep{bhowmick-jana-2021-sentiment,wahid2019cricket, Hoq2021SentimentAO}.

The ongoing shift in ICL with LLMs highlights the value of retrieval augmentation, where sourcing instructions with semantic information has become pivotal~\citep{shi2023replug}. In the multilingual ICL domain, methodologies like MEGA~\citep{ahuja2023mega} often narrow their focus to task-centric instructions, lacking in-depth semantic insights due to their reliance on random prompt selection. Complementing these advances, the work on CORAC~\citep{asai2021question} represents a significant leap forward in multilingual QA models, demonstrating a many-to-many approach that avoids language-specific data and retrieval modules, which is particularly beneficial for low-resource languages.In contrast, strategies like PARC~\citep{nie-etal-2023-cross} propose a more comprehensive methodology by obtaining semantically aligned instructions from high-resource languages.

Building on these methods, our work introduces novel perspectives and aims to bridge gaps. While MEGA provides task-centric instructions, we integrate deeper semantic understanding. We embrace a cross-lingual approach akin to PARC, as depicted in Figure~\ref{fig:parc}. In contrast to PARC's focus on masked language models like mBERT and XLMR, we explore the potential of larger, decoder-only multilingual pretrained language models (MPLMs) - BLOOM and BLOOMZ. Our focus is on addressing both classification and generation tasks in a cohesive generative style, emphasizing instruction execution~\citep{muennighoff-etal-2023-crosslingual, Scao2022BLOOMA1}.

This paper explores the application of cross-lingual retrieval-augmented ICL (CREA-ICL) to a specific low-resourced case, the Bangla language-covering text classification and generation tasks. We prioritize the effective execution of instructions. Our main contributions are:
\begin{itemize}
\item A comprehensive evaluation of cross-lingual retrieval augmented ICL, highlighting consistent improvements over MPLMs' zero-shot performance on Bangla classification tasks.
\item An in-depth analysis revealing the challenges in Bangla generation task, providing insights into the performance dynamics in both the classification and generation domains.
\item A pioneering exploration to adapt PARC for generative models, BLOOM and BLOOMZ, providing insights for a unified pipeline of CREA-ICL.
\end{itemize}

\section{Related Work}




\paragraph{Retrieval Augmented Prompt}
External knowledge extracted by information retrieval is often leveraged to solve NLP tasks. Two types of representations have been used for retrieval: (1) sparse bag-of-words representations \citep{chen-etal-2017-reading, Wang2018R3RR}, and (2) dense representation learned by neural networks \citep{Qu2020RocketQAAO}. Dense representations come either from contextual token embeddings \citep{May2019OnMS, Zhang2020BERTScoreET} or from sentence encoders \citep{conneau-etal-2017-supervised, cer2018universal}. \citet{reimers-2019-sentence-bert} propose sentence transformers to create semantically meaningful sentence embeddings by applying siamese and triplet network structures to transformer-based pretrained language models. By using knowledge distillation, sentence transformers can be expanded to support various languages as multilingual sentence transformers \citep{reimers-2020-multilingual-sentence-bert}, allowing for cross-lingual retrieval.

\citet{brown2020language} has shown that LLMs like GPT-3 can acquire task-solving abilities by incorporating input-output pairs as context. The in-context learning approach involves concatenating input with randomly selected examples from the training dataset, which is also called the prompting method. 
Recent research \citep{gao-etal-2021-making, liu-etal-2022-makes, liu-etal-2023-semantic, shi2023replug} has extended this idea by improving prompts for pre-trained models by incorporating semantically similar examples. They apply the retrieval augmented method to discrete prompts, which are represented by tokens instead of vectors in a continuous space.
They use them either for finetuning in few-shot settings or for zero-shot learning. \citet{Chowdhury2022NoveltyCP} use a similar kNN-based retrieval method for tuning the soft prompts in a continuous space with a standard supervised training setup.

\paragraph{Multilingual In-Context Learning}

The effectiveness of prompting methods for English models extends to multilingual models in cross-lingual transfer learning as well. \citet{zhao-schutze-2021-discrete} and \citet{huang-etal-2022-zero} investigated the prompt-based learning with multilingual PLMs.~\citet{nie-etal-2023-cross} incorporated augmented the prompt with cross-lingual retrieval samples in the multilingual understanding and proposed the PARC pipeline. PARC enhances the zero-shot learning performance for low-resource languages by cross-lingual retrieval from labeled or unlabeled high-resource languages. In the PARC pipeline, the cross-lingual retrieval first uses an low-resource language input sample as a query to find the semantically most similar high-resource language sample in the corpus. The recovered sample's label is received either from the corpus (labeled setting) or by self-prediction (unlabeled setting). The retrieved HRL sample together with its label, and the input sample are reformulated as prompts. Concatenation is used to generate the cross-lingual retrieval-augmented prompt, which is then used by the multilingual PLM to make a prediction.
\citet{tanwar-etal-2023-multilingual} augmented the prompt with not only cross-lingual semantic information but also additional task information. However, previous studies mainly concentrated on the multilingual encoder or encoder-decoder models, while our work extends the PARC pipeline to the decoder-only multilingual LLMs.

\paragraph{Multilingual LLMs}
In the era of LLMs, BLOOMZ and mT0~\citep{muennighoff-etal-2023-crosslingual} are two representative newly emerging multilingual models. These two multilingual LLMs are fine-tuned on xP3, a multilingual multitask fine-tuning dataset, and based on the pre-trained models BLOOM~\citep{Scao2022BLOOMA1} and mT5~\citep{xue-etal-2021-mt5}, respectively. Six different sizes of BLOOMZ models are released from 560M to 176B and 5 different sizes of mT0 models are released from 300M to 13B. These multilingual LLMs open up the possibility for conducting few- and zero-shot cross-lingual in-context learning, as demonstrated by recent benchmarking efforts, for example, MEGA~\citep{ahuja2023mega} and BUFFET~\citep{asai2023buffet}.

\section{Methodology}
Our research extends the work of \citet{nie-etal-2023-cross} by focusing on improving multilingual pre-trained language models (MPLMs) for low-resource languages in a zero-shot setting. Figure~\ref{fig:parc} illustrates our two-stage pipeline. At its core, this pipeline synergistically combines the strengths of MPLMs with the semantic depth of high-resource languages for both classification and generation tasks.

\subsection{Cross-Lingual Retrieval}
The foundation of our method lies in the efficient retrieval of semantically relevant samples from high-resource languages, given a low-resource language input. Formally, for a given input sentence $q$  (the input test for classification and summarization task and the question for QA task) from a low-resource language, the cross-lingual retriever maps it to a vector $q_{embed}$ in a shared embedding space using a function $Embed$:
$$
q_{embed} = Embed(q)
$$
For each document \( d_i \) in the high-resource language corpus, we compute its cosine similarity with \( q_{embed} \):
\[
Sim(q_{embed}, d_i) = \cos(q_{embed}, d_i)
\]
The top \(k\) documents, which are most semantically aligned to the input, are then retrieved:
\[
R_{indices} = \arg\max_{i \in \{1, \ldots, |d|\}}^k Sim(q_{embed}, d_i)
\]
\[
R = \{ d_i | i \in R_{indices} \}
\]
In cases where the retrieved documents \( d_i \) are unlabeled, a self-prediction mechanism that feeds \( d_i \) to the MPLM provides the necessary annotations.

\subsection{Prompt Engineering}

Using the semantically-rich retrieved samples \( R \) and the original input \( q \), we craft a contextually-enriched input \( \hat{q} \) using a predefined prompt template \( P \):
\[
\hat{q} = P(q, R)
\]
This template \( P \) not only encapsulates the task-specific instructions but adeptly combines \( q \) and \( R \) to maximize the model's comprehension. The transformed input \( \hat{q} \) is then processed by the MPLM to produce the desired output. 

Depending on the architecture of the chosen MPLM, for autoregressive models like GPT variants, the model naturally generates an output sequence \( \mathbf{y} \) for a given input \( \hat{q} \):
\[
\mathbf{y} = MPLM(\hat{q})
\], whereas encoder models leverage a mask token prediction mechanism, utilizing a \textit{verbalizer} to map labels to their corresponding linguistic representations, which will be shown in the experiment settings.


\section{Experiments}
To empirically validate our methodology, we design experiments that include both classification and generation. These experiments provide insights into the effectiveness of integrating MPLMs with high-resource language semantics, to improve ICL for low-resource languages.

\subsection{Tasks and Datasets}
\paragraph{Vio-Lens} The Vio-Lens dataset~\citep{SahaAndJunaed} provides a rich collection of YouTube comments related to violent episodes in the Bengal region, structured for classification. Our prompt template \( P \) includes:
\begin{itemize}
\item Autoregressive models: \\
 \texttt{"Reflecting on the statement \{text\}, which aggressive level does it resonate with: non-aggressive, slightly aggressive, or highly aggressive?"}
\item Mask prediction models:
 \texttt{"The underlying theme in \{text\} is {[}MASK{]}."} \\
 with the \textit{verbalizer}:\\
 $v(0)= $ \texttt{"assaultive"}, 
 $v(1) = $ \texttt{"indirect"}, 
 $v(2)= $ \texttt{"peaceful"}
\end{itemize}
As the retrieval corpora for Vio-Lens, we resort to the ETHOS dataset~\citep{mollas2020ethos}, a comprehensive repository targeting online hate speech detection as the retrieval sentence pool. This repository provides a dataset designed to identify hate speech on social media, which contains 998 comments, each labeled for the presence or absence of hate speech. Since the labels are inconsistent, we rely on self-prediction to annotate the labels.

\paragraph{SentNoB} 
The SentNoB dataset~\citep{islam-etal-2021-sentnob-dataset} is crafted to dissect the sentiment embedded within Bangla texts. Our prompt template \( P \) is articulated as:
\begin{itemize}
\item Autoregressive models: \\
\texttt{"Text: \{text\} What is a possible sentiment for the text given the following options?"}
\item Mask prediction models:
\texttt{"\{text\} Sentiment: {[}MASK{]}"}  \\
with the \textit{verbalizer}:\\
$v(0) = $ \texttt{positive},  
$v(1) = $ \texttt{neural}, 
$v(2) = $ \texttt{negative}
\end{itemize}
For SentNoB, we use the labeled training set of English Sentiment Analysis dataset~\citep{rosenthal-etal-2017-semeval}, which consists of tweets annotated for sentiment on 2-, 3-, and 5-point scales with labels positive, negative, and neutral.

\paragraph{XLSum}
As a typical representative generation task, XLSum~\citep{hasan-etal-2021-xl}, the multilingual text summarization dataset consisting of 1.35 million pairs of articles and their corresponding summaries. These pairs have been expertly annotated by the BBC and meticulously extracted through a series of carefully designed heuristic methods. We evaluate its Bangla subset to assess the performance of generation. The English training set is used as the retrieval corpus. 
The prompt template is defined as follows:

 \texttt{"\{text\} Generate a concise summary of the above text using the same language as the original text:"}

\paragraph{XQuAD}
the XQuAD (Cross-lingual Question Answering Dataset)~\citep{Artetxe2019OnTC} was used, which contains topic-diverse, manually-curated question-answer pairs with their respective contexts in 11 languages. Greek and Romanian language subsets are being evaluated. These pairs have been carefully translated from the English SQuAD v1.1 dataset to ensure high quality translations. The prompt template is defined as follows:

 \texttt{"context: \{context\} question: \{question\} answer:"}

\subsection{Models}
\paragraph{BLOOM} is an autoregressive Large Language Model trained on a diverse corpus to generate text based on prompts~\citep{Scao2022BLOOMA1}. It is capable of generating coherent text in 46 languages. 

\paragraph{BLOOMZ} takes a novel approach in the MPLM landscape by applying Bloom filters in the context of language models~\citep{muennighoff-etal-2023-crosslingual}. This allows the model to use high-resource languages to improve embeddings for low-resource languages, effectively bridging the gap between languages with different levels of available resources. 

\paragraph{mBERT} is an early MPLM that extends the original BERT model~\citep{DBLP:journals/corr/abs-1810-04805}. It is pre-trained on a corpus of 104 languages, using shared WordPiece vocabularies and a unified architecture for all languages.

\paragraph{mT5} or Multilingual T5~\citep{xue-etal-2021-mt5}, is an extension of the T5 (Text-to-Text Transfer Transformer) model~\citep{raffel2020exploring} specifically designed for multilingual capabilities. Pre-trained on mC4, a large multilingual dataset, mT5 demonstrates multilingual capabilities by transforming input text sequences into output sequences.

\paragraph{Cross-Lingual Retriever} We followed  \citet{nie-etal-2023-cross} to use 
the multilingual sentence transformer ``\textit{paraphrase-multilingual-mpnet-base-v2}''~\citep{reimers-gurevych-2019-sentence}. This transformer maps sentences and paragraphs into a 768-dimensional dense vector space. Such a high-dimensional embedding facilitates tasks such as clustering and semantic search.
Retrieval sample settings \(k\) are meticulously set at 1 and 3 for classification, and confined to 1 for summarization, 3 for QA.

\section{Results}
\subsection{Results of classification tasks}

\begin{table}[]
\footnotesize
\centering \begin{tabular}{lccc|lccc}
\toprule
Vio-Lens  & zero shot & k=1  & k=3  & SentNoB   & zero shot & k=1  & k=3 \\ 
\midrule
bloomz-3b & 0.19      & 0.2  & 0.24 & bloomz-3b & 0.34      & 0.44 & 0.44 \\ 
bloom-3b  & 0.00      & 0.00 & 0.00 & bloom-3b  & 0.00      & 0.00 & 0.00 \\ 
mbert     & 0.21      & 0.28 & 0.29 & mbert     & 0.30      & 0.36 & 0.37 \\ 
\bottomrule
\end{tabular}
\caption{F1-scores of the two classification tasks: Bangla zero-shot baseline 
and our main method CREA-ICL with $k$ retrieval augmented prompts.}
\label{tab:results of classification test}
\end{table}

\begin{table*}[]
\scriptsize
\centering \begin{tabular}{lccccccccc}
\toprule
\multicolumn{1}{l}{}                 
& \multicolumn{3}{c}{zero shot}                        
& \multicolumn{3}{c}{k=1}                          
& \multicolumn{3}{c}{k=3}      
\\ \midrule
\multicolumn{1}{l}{bloomz-3b}                 
& \multicolumn{1}{c}{precision} 
& \multicolumn{1}{c}{recall}
& \multicolumn{1}{c}{f1-score}   
& \multicolumn{1}{c}{precision}
& \multicolumn{1}{c}{recall} 
& \multicolumn{1}{c}{f1-score}
& \multicolumn{1}{c}{precision}
& \multicolumn{1}{c}{recall} 
& \multicolumn{1}{c}{f1-score}    
\\ \midrule
\multicolumn{1}{l}{non-violence}     
& \multicolumn{1}{c}{0.00}      
& \multicolumn{1}{c}{0.00}   
& \multicolumn{1}{c}{0.00}    
& \multicolumn{1}{c}{0.00}     
& \multicolumn{1}{c}{0.00} 
& \multicolumn{1}{c}{0.00}   
& \multicolumn{1}{c}{0.50}    
& \multicolumn{1}{c}{0.04}  
& \multicolumn{1}{c}{0.08}
\\ 
\multicolumn{1}{l}{passive violence} 
& \multicolumn{1}{c}{0.36}     
& \multicolumn{1}{c}{0.89}  
& \multicolumn{1}{c}{0.51} 
& \multicolumn{1}{c}{0.36} 
& \multicolumn{1}{c}{0.97} 
& \multicolumn{1}{c}{0.52}
& \multicolumn{1}{c}{0.36}
& \multicolumn{1}{c}{0.91}  
& \multicolumn{1}{c}{0.51} 
\\ 
\multicolumn{1}{l}{direct violence} 
& \multicolumn{1}{c}{0.09}   
& \multicolumn{1}{c}{0.10}   
& \multicolumn{1}{c}{0.10}  
& \multicolumn{1}{c}{0.18}  
& \multicolumn{1}{c}{0.06}  
& \multicolumn{1}{c}{0.09}  
& \multicolumn{1}{c}{0.17} 
& \multicolumn{1}{c}{0.07}  
& \multicolumn{1}{c}{0.10}
\\ 
\multicolumn{1}{l}{accuracy}        
& \multicolumn{1}{l}{}         
& \multicolumn{1}{l}{}      
& \multicolumn{1}{c}{{0.33}}
& \multicolumn{1}{l}{}      
& \multicolumn{1}{l}{}       
& \multicolumn{1}{c}{0.35} 
& \multicolumn{1}{l}{}     
& \multicolumn{1}{l}{}      
& \multicolumn{1}{c}{\textbf{0.36}} 
\\ 
\multicolumn{1}{l}{macro avg}      
& \multicolumn{1}{c}{0.15}     
& \multicolumn{1}{c}{0.33} 
& \multicolumn{1}{c}{\textbf{0.20}}    
& \multicolumn{1}{c}{0.18}  
& \multicolumn{1}{c}{0.34} 
& \multicolumn{1}{c}{\textbf{0.20}} 
& \multicolumn{1}{c}{0.26} 
& \multicolumn{1}{c}{0.26}  
& \multicolumn{1}{c}{{0.17}}
\\ 
\multicolumn{1}{l}{weighted avg}    
& \multicolumn{1}{c}{0.14}      
& \multicolumn{1}{c}{0.33}  
& \multicolumn{1}{c}{{0.19}}
& \multicolumn{1}{c}{0.15}    
& \multicolumn{1}{c}{0.35}  
& \multicolumn{1}{c}{{0.20}}   
& \multicolumn{1}{c}{0.42}    
& \multicolumn{1}{c}{0.36}  
& \multicolumn{1}{c}{\textbf{0.24}} 
\\ \toprule
\multicolumn{1}{l}{mbert}                 
& \multicolumn{1}{c}{precision} 
& \multicolumn{1}{c}{recall}
& \multicolumn{1}{c}{f1-score}   
& \multicolumn{1}{c}{precision}
& \multicolumn{1}{c}{recall} 
& \multicolumn{1}{c}{f1-score}
& \multicolumn{1}{c}{precision}
& \multicolumn{1}{c}{recall} 
& \multicolumn{1}{c}{f1-score}    
\\ \midrule                       
\multicolumn{1}{l}{non}     
& \multicolumn{1}{c}{0.46}    
& \multicolumn{1}{c}{0.28}  
& \multicolumn{1}{c}{0.35}  
& \multicolumn{1}{c}{0.47} 
& \multicolumn{1}{c}{0.52} 
& \multicolumn{1}{c}{0.49} 
& \multicolumn{1}{c}{0.46} 
& \multicolumn{1}{c}{0.55}  
& \multicolumn{1}{c}{0.50} 
\\ 
\multicolumn{1}{l}{passive}    
& \multicolumn{1}{c}{0.38}     
& \multicolumn{1}{c}{0.01}  
& \multicolumn{1}{c}{0.02}  
& \multicolumn{1}{c}{1.00} 
& \multicolumn{1}{c}{0.00}  
& \multicolumn{1}{c}{0.00}  
& \multicolumn{1}{c}{0.00}  
& \multicolumn{1}{c}{0.00}   
& \multicolumn{1}{c}{0.00} 
\\ 
\multicolumn{1}{l}{direct}   
& \multicolumn{1}{c}{0.09}  
& \multicolumn{1}{c}{0.60}  
& \multicolumn{1}{c}{0.16}  
& \multicolumn{1}{c}{0.09} 
& \multicolumn{1}{c}{0.36} 
& \multicolumn{1}{c}{0.14} 
& \multicolumn{1}{c}{0.08}
& \multicolumn{1}{c}{0.30}   
& \multicolumn{1}{c}{0.13}
\\ 
\multicolumn{1}{l}{accuracy}     
& \multicolumn{1}{l}{}      
& \multicolumn{1}{l}{}     
& \multicolumn{1}{c}{0.22} 
& \multicolumn{1}{l}{}      
& \multicolumn{1}{l}{}      
& \multicolumn{1}{c}{0.32} 
& \multicolumn{1}{l}{}      
& \multicolumn{1}{l}{}      
& \multicolumn{1}{c}{\textbf{0.33}}
\\ 
\multicolumn{1}{l}{macro avg} 
& \multicolumn{1}{c}{0.31}   
& \multicolumn{1}{c}{0.30}  
& \multicolumn{1}{c}{0.18}   
& \multicolumn{1}{c}{0.52} 
& \multicolumn{1}{c}{0.29} 
& \multicolumn{1}{c}{\textbf{0.21}} 
& \multicolumn{1}{c}{0.18} 
& \multicolumn{1}{c}{0.28}  
& \multicolumn{1}{c}{\textbf{0.21}}
\\ 
\multicolumn{1}{l}{weighted avg}  
& \multicolumn{1}{c}{0.40}    
& \multicolumn{1}{c}{0.22}
& \multicolumn{1}{c}{0.21} 
& \multicolumn{1}{c}{0.62} 
& \multicolumn{1}{c}{0.32} 
& \multicolumn{1}{c}{0.28}
& \multicolumn{1}{c}{0.26} 
& \multicolumn{1}{c}{0.33}  
& \multicolumn{1}{c}{\textbf{0.29}} 
\\ \bottomrule
\end{tabular}
\\
\begin{tabular}{lccccccccc}
\toprule
\multicolumn{1}{l}{}             
& \multicolumn{3}{c}{zero shot} 
& \multicolumn{3}{c}{k=1}
& \multicolumn{3}{c}{k=3}\\
\midrule
\multicolumn{1}{l}{bloomz-3b}             
& \multicolumn{1}{c}{precision} 
& \multicolumn{1}{c}{recall} 
& \multicolumn{1}{c}{f1-score}      
& \multicolumn{1}{c}{precision} 
& \multicolumn{1}{c}{recall} 
& \multicolumn{1}{c}{f1-score} 
& \multicolumn{1}{c}{precision} 
& \multicolumn{1}{c}{recall} 
& \multicolumn{1}{c}{f1-score} 
\\ 
\midrule
\multicolumn{1}{l}{Negative}     
& \multicolumn{1}{c}{0.66}      
& \multicolumn{1}{c}{0.78}   
& \multicolumn{1}{c}{0.71}          
& \multicolumn{1}{c}{0.61}      
& \multicolumn{1}{c}{0.85}   
& \multicolumn{1}{c}{0.71}     
& \multicolumn{1}{c}{0.62}      
& \multicolumn{1}{c}{0.86}   
& \multicolumn{1}{c}{0.72}
\\ 
\multicolumn{1}{l}{Neutral}      
& \multicolumn{1}{c}{0.00}      
& \multicolumn{1}{c}{0.00}   
& \multicolumn{1}{c}{0.00} 
& \multicolumn{1}{c}{0.23}      
& \multicolumn{1}{c}{0.02}   
& \multicolumn{1}{c}{0.03}
& \multicolumn{1}{c}{0.18}      
& \multicolumn{1}{c}{0.01}   
& \multicolumn{1}{c}{0.01}  \\ 
\multicolumn{1}{l}{Positive}     
& \multicolumn{1}{c}{0.57}      
& \multicolumn{1}{c}{0.72}   
& \multicolumn{1}{c}{0.64}
& \multicolumn{1}{c}{0.60}      
& \multicolumn{1}{c}{0.56}   
& \multicolumn{1}{c}{0.58}
& \multicolumn{1}{c}{0.59}      
& \multicolumn{1}{c}{0.57}   
& \multicolumn{1}{c}{0.58}\\ 
\multicolumn{1}{l}{accuracy}     
& \multicolumn{1}{l}{}          
& \multicolumn{1}{l}{}       
& \multicolumn{1}{c}{\textbf{0.61}}
& \multicolumn{1}{l}{}          
& \multicolumn{1}{l}{}       
& \multicolumn{1}{c}{0.60}     
& \multicolumn{1}{l}{}          
& \multicolumn{1}{l}{}       
& \multicolumn{1}{c}{\textbf{0.61}}\\ 
\multicolumn{1}{l}{macro avg}    
& \multicolumn{1}{c}{0.31}      
& \multicolumn{1}{c}{0.37}   
& \multicolumn{1}{c}{0.34}          
& \multicolumn{1}{c}{0.48}      
& \multicolumn{1}{c}{0.48}   
& \multicolumn{1}{c}{\textbf{0.44}}    
& \multicolumn{1}{c}{0.47}      
& \multicolumn{1}{c}{0.48}   
& \multicolumn{1}{c}{\textbf{0.44}} \\
\multicolumn{1}{l}{weighted avg} 
& \multicolumn{1}{c}{0.51}      
& \multicolumn{1}{c}{0.61}   
& \multicolumn{1}{c}{\textbf{0.55}} 
& \multicolumn{1}{c}{0.53}      
& \multicolumn{1}{c}{0.60}   
& \multicolumn{1}{c}{0.54}    
& \multicolumn{1}{c}{0.53}     
& \multicolumn{1}{c}{0.61}  
& \multicolumn{1}{c}{0.54}\\ 
\midrule
\multicolumn{1}{l}{mbert}             
& \multicolumn{1}{c}{precision} 
& \multicolumn{1}{c}{recall} 
& \multicolumn{1}{c}{f1-score}
& \multicolumn{1}{c}{precision} 
& \multicolumn{1}{c}{recall} 
& \multicolumn{1}{c}{f1-score} 
& \multicolumn{1}{c}{precision} 
& \multicolumn{1}{c}{recall} 
& \multicolumn{1}{c}{f1-score} \\ 
\midrule
\multicolumn{1}{l}{negative}     
& \multicolumn{1}{c}{0.55}      
& \multicolumn{1}{c}{0.48}  
& \multicolumn{1}{c}{0.51}  
& \multicolumn{1}{c}{0.60}  
& \multicolumn{1}{c}{0.38}  
& \multicolumn{1}{c}{0.47}    
& \multicolumn{1}{c}{0.60}  
& \multicolumn{1}{c}{0.45}   
& \multicolumn{1}{c}{0.52}  
\\
\multicolumn{1}{l}{neutral}     
& \multicolumn{1}{c}{0.18}  
& \multicolumn{1}{c}{0.47}   
& \multicolumn{1}{c}{0.26}  
& \multicolumn{1}{c}{0.19}  
& \multicolumn{1}{c}{0.46}  
& \multicolumn{1}{c}{0.27}  
& \multicolumn{1}{c}{0.20}  
& \multicolumn{1}{c}{0.49}  
& \multicolumn{1}{c}{0.28}     
\\ 
\multicolumn{1}{l}{positive}     
& \multicolumn{1}{c}{0.40}      
& \multicolumn{1}{c}{0.07}  
& \multicolumn{1}{c}{0.13}   
& \multicolumn{1}{c}{0.43}  
& \multicolumn{1}{c}{0.29}  
& \multicolumn{1}{c}{0.35}  
& \multicolumn{1}{c}{0.46}  
& \multicolumn{1}{c}{0.24} 
& \multicolumn{1}{c}{0.32}   
\\ 
\multicolumn{1}{l}{accuracy}     
& \multicolumn{1}{l}{}        
& \multicolumn{1}{l}{}     
& \multicolumn{1}{c}{0.35} 
& \multicolumn{1}{l}{}    
& \multicolumn{1}{l}{}     
& \multicolumn{1}{c}{0.37} 
& \multicolumn{1}{l}{}     
& \multicolumn{1}{l}{}     
& \multicolumn{1}{c}{\textbf{0.39}}
\\ 
\multicolumn{1}{l}{macro avg}  
& \multicolumn{1}{c}{0.38}  
& \multicolumn{1}{c}{0.34} 
& \multicolumn{1}{c}{0.30} 
& \multicolumn{1}{c}{0.40} 
& \multicolumn{1}{c}{0.38} 
& \multicolumn{1}{c}{0.36} 
& \multicolumn{1}{c}{0.42} 
& \multicolumn{1}{c}{0.39}  
& \multicolumn{1}{c}{\textbf{0.37}}
\\ 
\multicolumn{1}{l}{weighted avg}
& \multicolumn{1}{c}{0.43}  
& \multicolumn{1}{c}{0.35}  
& \multicolumn{1}{c}{0.34}  
& \multicolumn{1}{c}{0.47}  
& \multicolumn{1}{c}{0.37}
& \multicolumn{1}{c}{0.39}
& \multicolumn{1}{c}{0.48} 
& \multicolumn{1}{c}{0.39}  
& \multicolumn{1}{c}{\textbf{0.41}}
\\ 
\bottomrule
\end{tabular}

\caption{Confusion matrix of CREA-ICL method in Vio-Lens (top) and SentNoB (bottom) test set of BLOOMZ-3b and mBERT.}
\label{tab:confusion matrix of main method in Vio-Lens test of bloomz-3b and mbert}
\end{table*}

Table~\ref{tab:results of classification test} provides a snapshot of the classification results. When we leveraged $k=3$ retrieved English prompts, Bloomz-3b demonstrated an improvement in F1-scores for both tasks by 5\% and 10\% respectively. The striking null results from Bloom-3b, in contrast to Bloomz-3b, emphasize the pivotal role instruction tuning plays in retrieval-augmented in-context learning. In comparison, the traditional masked MLM, mBERT, also registered an improvement of 8\% and 7\%, respectively.

A deeper dive into the confusion matrix, as presented in Table~\ref{tab:confusion matrix of main method in Vio-Lens test of bloomz-3b and mbert}, reveals intriguing insights:

1)  With a general assessment across micro, macro, and weighted F1 scores,  Bloomz-3b and mBERT gained improvement from the retrieval prompts.

2) Comparing the two models, Bloomz-3b's zero-shot setting tends to misclassify ``non-violence'' and ``Neutral'', and has a reduced macro F1 compared to its weighted F1, while mBERT has a more balanced distribution of confusion between ``non-violence'' (``Neutral'') and the other classes.

These may indicate that for classification tasks, the generative models struggle more with minority classes compared to masked prediction.

\subsection{Results of generation task}

\begin{table*}[]
\footnotesize
\centering \begin{tabular}{cc|ccc|ccc}
\toprule
\multicolumn{2}{c|}{LEAD-64}  & \multicolumn{3}{c|}{zero shot}    & \multicolumn{3}{c}{k=1}            \\
\multicolumn{1}{l}{} &         & mt5-base & bloomz-1b1 & bloomz-3b & mt5-base & bloomz-1b1 & bloomz-3b \\ 
\midrule
R-1                  & 18.17   & 5.01     & 22.08      & 22.36     & 0.97     & 10.84      & 6.61      \\
R-2                  & 5.23    & 0.84     & 7.11       & 7.88      & 0.13     & 2.80       & 1.52      \\
R-L                  & 12.73   & 4.83     & 18.43      & 18.60     & 0.91     & 9.11       & 5.56      \\
R-LSum               & 12.74   & 4.84     & 18.44      & 18.58     & 0.92     & 9.12       & 5.55      \\ 
\bottomrule
\end{tabular}
\\
\begin{tabular}{cccccc}
\toprule
         &    & \multicolumn{2}{c}{zero-shot} & \multicolumn{2}{c}{k=3} \\
         &    & EM            & F1            & EM         & F1         \\ \midrule
Greek    & el & 15.29      & 19.92          & 8.07          & 10.34      \\
Romanian & ro & 38.66      & 49.39          & 15.88         & 20.93      \\ \bottomrule
\end{tabular}
\caption{Results of Bangla summarization (top) and QA task (bottom), including the zero-shot baseline and CREA-ICL method with k=1 or k=3 retrieved samples.}
\label{tab:results of qa task}
\end{table*}

The summarization task results, as shown in Table~\ref{tab:results of qa task}, provide a comprehensive understanding of the models' capabilities:

\paragraph{LEAD-64} The LEAD-64 baseline operates on a straightforward extractive approach where the first 64 tokens of the input text are considered as its summary. Its commendable performance across various metrics reiterates the often-overlooked significance of the initial segments in articles or documents. These segments frequently encapsulate key points, making them effective summaries. In a zero-shot setting, LEAD-64's results surpass those of the mt5-base model. However, when compared to the Bloomz variants, it finds itself overshadowed. This result emphasizes the effect of instruction tuning that enhances ICL performance.

\paragraph{Zero-Shot Baseline} mt5-base's discernible underperformance across the board underscores its struggles in generating high-quality summaries without specialized domain adaptation or data enrichment. In stark contrast, the Bloomz models exhibit remarkable competency, with Bloomz-3b slightly outdoing Bloomz-1b1, especially in the context of the R-2 metric, which evaluates bigram comprehension.

\paragraph{CREA-ICL with k=1}
Retrieval augmentation seems to drastically affect the performance of mt5-base, reducing its score considerably. This could be due to noise introduced by the retrieved sample or ineffective use of the additional information.
For the Bloomz models, Bloomz-1b1 still retains decent performance, although there's a drop when compared to its zero-shot performance. Surprisingly, Blommz-3b shows a sharper drop, suggesting that the additional retrieval data may be more of a distraction than an advantage for this model configuration in the summarization task.


From the table \ref{tab:results of qa task}, which shows the performance metrics of the QA task in Greek and Romanian. The CREA-ICL method with k=3 using the Bloomz-3b model seems to be less effective than the zero-shot baseline using the same model for both languages. Both EM and F1-score metrics demonstrate a significant performance advantage, nearly doubling their values compared to CREA-ICL.

\begin{figure*}
    \centering
    \includegraphics[width=0.49\linewidth]{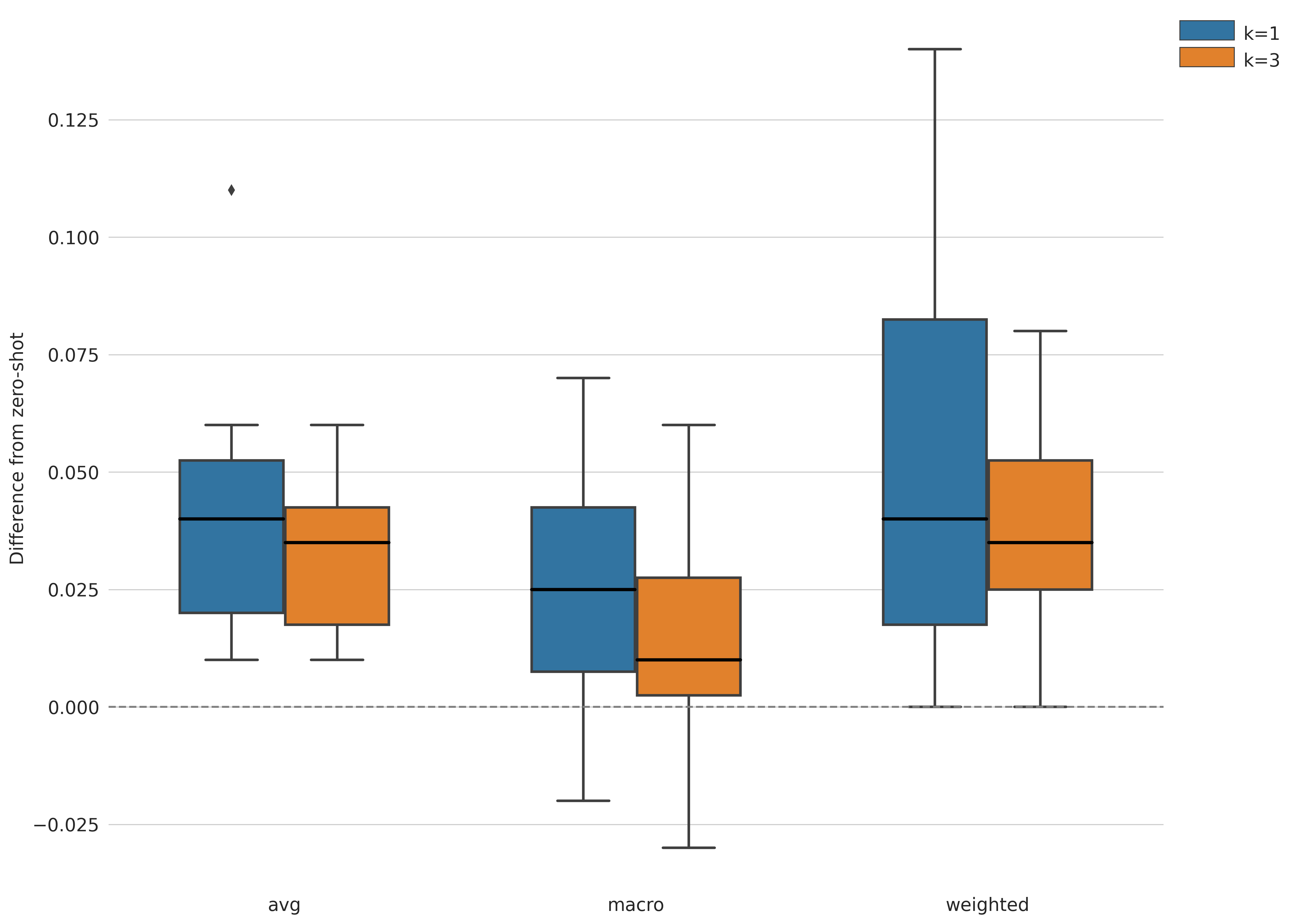}
    \includegraphics[width=0.49\linewidth]{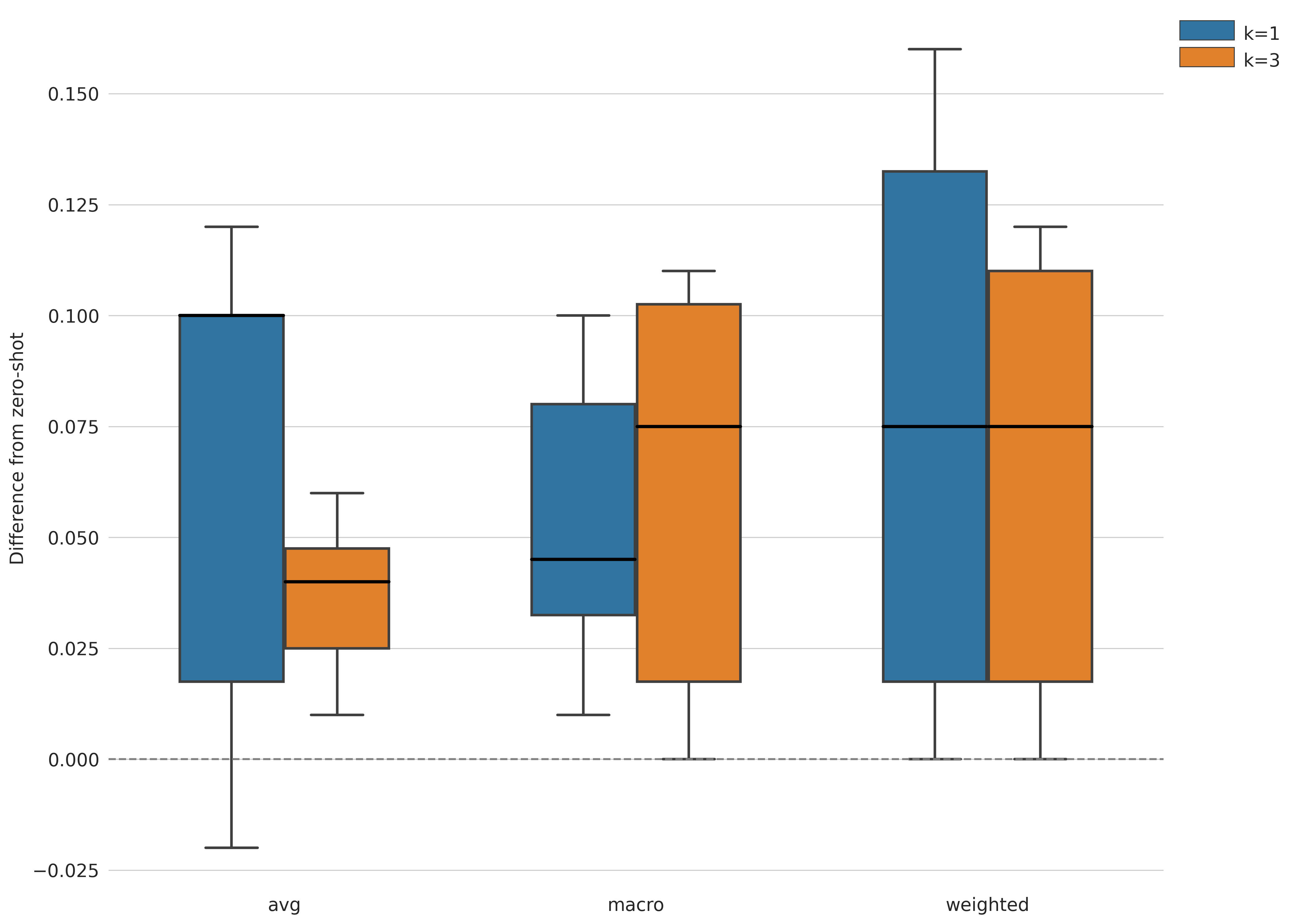}
    \caption{Model performance over differences between zero-shot and CREA-ICL method with k=1 and k=3 demonstrations for Vio-Lens test set using bloomz-3b (left) and mbert (right).}
    \label{fig:boxplot_task1_test_bloomz3b}
\end{figure*}

\subsection{Analysis and Discussion}

When analyzing the effectiveness of different models across different tasks, it becomes clear that for classification tasks, advanced models have the ability to identify complex categories. This ability is not limited to a single language or dataset, but extends to a variety of linguistic contexts. Zero-shot learning, a method in which models apply knowledge without direct task-specific training, proves to be versatile and shows remarkable capability even without task-specific fine-tuning. Exploration of different numbers of retrieved samples (\(k\)) shows that increasing \(k\) does not necessarily improve performance, suggesting strategic information filtering and noise reduction capabilities of the models.

In summarization tasks, maintaining coherence and relevance is critical. In such generative tasks, models designed for generation generally outperform those designed for extraction in the absence of task-specific examples. However, these generative models are sensitive to the addition of retrieved information, which can affect their performance. The balance between performance and computational effort becomes clear when analyzing resource allocation, and shows that while models such as Bloomz-3b demonstrate superior performance, larger models do not guarantee better results across all evaluation criteria. This suggests the need for a more selective approach to model selection.

Furthermore, in the question-answering (QA) domain, zero-shot methods have been shown to outperform the CREA-ICL approach. This underscores the robustness of zero-shot strategies in classification contexts, but also points to their limitations in generative tasks, where they may not always be the optimal choice. This finding encourages a more nuanced application of models, depending on the nature of the task at hand.



Our analysis of cross-lingual tasks shows that classification accuracy often has an advantage over that of generative tasks, especially when dealing with multilingual data. Classification tasks typically require the model to produce short, constrained outputs, often limited to a single word or short phrase from a predefined set of options. This specificity in response reduces the likelihood of encountering language confusion.

\begin{figure*}
	\centering
	\includegraphics[width=0.7\linewidth]{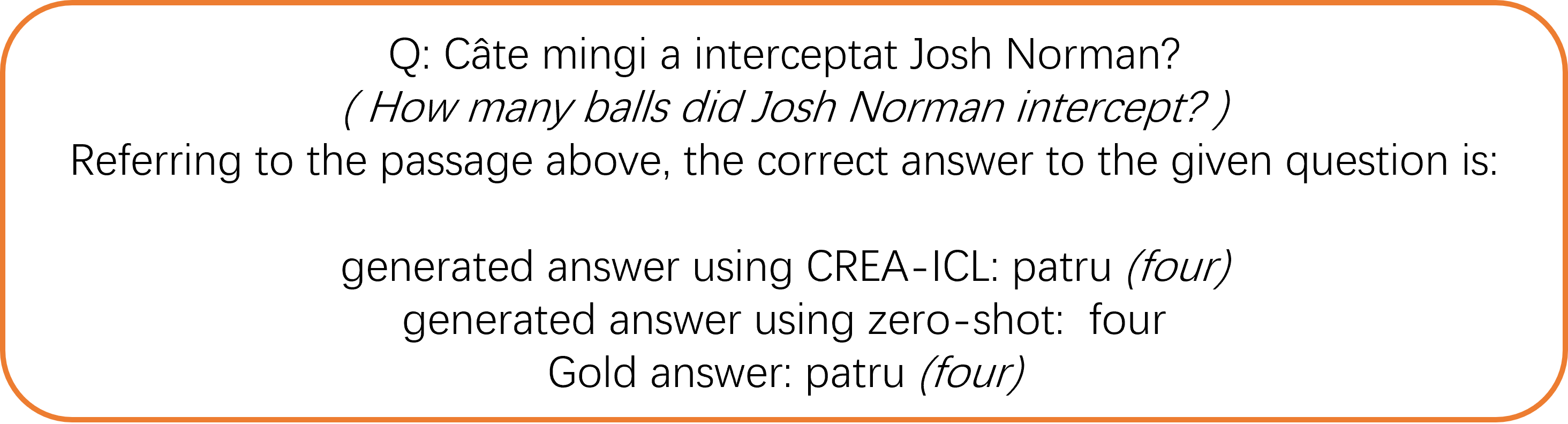}
	\caption{Example showing the different answer language when answering a Romanian question}
	\label{fig:wrong-number-Romanian}
\end{figure*}

Conversely, generative tasks require the production of longer sequences of text, which increases the challenge of maintaining language consistency. This is evident in scenarios where a model, while capturing the correct conceptual meaning, outputs in an unexpected language. Figure~\ref{fig:wrong-number-Romanian} illustrates this, where the generated output is conceptually accurate but linguistically misaligned. Our methodology, with its dependence on English templates, may inadvertently worsen this problem rather than mitigate it, as the templates may bias the generative process towards English.

Furthermore, the metrics currently used for generative tasks still struggle with evaluating output that is linguistically inconsistent. Our generative tasks have been modified to include a language constraint that requires the output to match the language of the original text. However, as shown in Table~\ref{tab:with and without language constraint}, this adjustment resulted in only marginal improvements. The implications are significant: when evaluating generative tasks in a cross-lingual context, there remains a significant challenge in accurately capturing the quality of language-specific output. In the future, the development of metrics that can better account for language consistency will be crucial for evaluating the true effectiveness of models in generative multilingual scenarios.

\begin{table*}[]
\footnotesize
\centering
\begin{tabular}{lcccc}
\toprule
                                                                  & R-1  & R-2  & R-L  & R-L-Sum \\ \hline
\{text\} Generate a concise summary of the given text             & 0.55 & 0.07 & 0.53 & 0.53    \\ \midrule
...using the same language as the original text(\{target\_lang\}) & 0.97 & 0.13 & 0.91 & 0.92    \\ \bottomrule
\end{tabular}
\caption{results with and without language constraint in prompt templates of summarization task using mt5-base model with k=1}
\label{tab:with and without language constraint}
\end{table*}


\section{Ablation Study}

\subsection{The Stability across Templates}

In our investigation of the Vio-Lens dataset, we evaluated the classification capabilities of Bloomz-3b and mBERT by categorizing text samples. Our goal was to compare the effectiveness of retrieval-augmented prompting to a zero-shot baseline by analyzing performance across different prompt templates.

We applied different templates  
to both Bloomz-3b and mBERT and generated a boxplot (Figure~\ref{fig:boxplot_task1_test_bloomz3b}) to show the F1 score variations when using the CREA-ICL method versus the zero-shot baseline. The visual representation confirmed that there was a consistent improvement in performance with the retrieval-augmented English prompts compared to the zero-shot baseline, which did not focus on any specific language. Interestingly, mBERT showed a more pronounced improvement in F1 scores than Bloomz-3b when moving from the zero-shot baseline to the retrieval-augmented prompts. This finding highlights the potential of retrieval augmentation to improve the model's text classification performance beyond the context of a single language.

\subsection{Evaluating the Influence of Non-English Prompt Templates}
\begin{table*}[]
\footnotesize
\centering \begin{tabular}{lccccccc}
\toprule
& \multicolumn{3}{c}{k=1}        
& \multicolumn{3}{c}{k=3}  
\\ 
& \multicolumn{1}{c}{precision} 
& \multicolumn{1}{c}{recall} 
& \multicolumn{1}{c}{f1-score} 
& \multicolumn{1}{c}{precision} 
& \multicolumn{1}{c}{recall} 
& \multicolumn{1}{c}{f1-score} 
\\ \midrule
bangla prompt & \multicolumn{7}{l}{
{\bng paThY}: \{text\} {\bng inm/nilikht ibkl/pguil ed{O}Ja paeThYr jnY sm/bhabY Anubhuuit kii?}
}\\ 

Negative      
& \multicolumn{1}{c}{0.58}     
& \multicolumn{1}{c}{0.20} 
& \multicolumn{1}{c}{0.29}  
& \multicolumn{1}{c}{0.60} 
& \multicolumn{1}{c}{0.59} 
& \multicolumn{1}{c}{0.60} 
\\ 
Neutral       
& \multicolumn{1}{c}{0.20}    
& \multicolumn{1}{c}{0.06}  
& \multicolumn{1}{c}{0.10}   
& \multicolumn{1}{c}{0.16}   
& \multicolumn{1}{c}{0.08}   
& \multicolumn{1}{c}{0.11}   
\\ 
Positive      
& \multicolumn{1}{c}{0.60}  
& \multicolumn{1}{c}{0.08}  
& \multicolumn{1}{c}{0.15}   
& \multicolumn{1}{c}{0.50} 
& \multicolumn{1}{c}{0.44} 
& \multicolumn{1}{c}{0.47} 
\\
accuracy    
& \multicolumn{1}{c}{} 
& \multicolumn{1}{c}{}  
& \multicolumn{1}{c}{0.14}  
& \multicolumn{1}{c}{}    
& \multicolumn{1}{c}{}      
& \multicolumn{1}{c}{0.45} 
\\
macro avg  
& \multicolumn{1}{c}{0.34}   
& \multicolumn{1}{c}{0.09}   
& \multicolumn{1}{c}{0.13}  
& \multicolumn{1}{c}{0.32} 
& \multicolumn{1}{c}{0.28} 
& \multicolumn{1}{c}{0.29}
\\
weighted avg  
& \multicolumn{1}{c}{0.51}  
& \multicolumn{1}{c}{0.14}
& \multicolumn{1}{c}{0.21}  
& \multicolumn{1}{c}{0.49} 
& \multicolumn{1}{c}{0.45} 
& \multicolumn{1}{c}{0.46} 
\\ \midrule
hindi prompt  & \multicolumn{7}{c}{
{\dn pAW}: \{text\} {\dn En\3DFwElEKt EvkSpo{\qva} ko d\?Kt\? \7{h}e pAW k\? Ele s\2BAEvt BAvnA \3C8wA h\4{\rs ?\re}}
}\\ 
Negative    
& \multicolumn{1}{c}{0.61}   
& \multicolumn{1}{c}{0.45}  
& \multicolumn{1}{c}{0.52}  
& \multicolumn{1}{c}{0.62}  
& \multicolumn{1}{c}{0.74}  
& \multicolumn{1}{c}{0.68}  
\\
Neutral      
& \multicolumn{1}{c}{0.18}  
& \multicolumn{1}{c}{0.34}   
& \multicolumn{1}{c}{0.24} 
& \multicolumn{1}{c}{0.19} 
& \multicolumn{1}{c}{0.14}  
& \multicolumn{1}{c}{0.16}  
\\
Positive    
& \multicolumn{1}{c}{0.55}    
& \multicolumn{1}{c}{0.31}   
& \multicolumn{1}{c}{0.40}  
& \multicolumn{1}{c}{0.57}  
& \multicolumn{1}{c}{0.48}  
& \multicolumn{1}{c}{0.52}  
\\
accuracy     
& \multicolumn{1}{c}{}       
& \multicolumn{1}{c}{}       
& \multicolumn{1}{c}{0.39}   
& \multicolumn{1}{c}{}      
& \multicolumn{1}{c}{}      
& \multicolumn{1}{c}{0.54}   
\\
macro avg   
& \multicolumn{1}{c}{0.34}    
& \multicolumn{1}{c}{0.28}  
& \multicolumn{1}{c}{0.29}  
& \multicolumn{1}{c}{0.34}  
& \multicolumn{1}{c}{0.34}  
& \multicolumn{1}{c}{0.34}  
\\
weighted avg
& \multicolumn{1}{c}{0.51}  
& \multicolumn{1}{c}{0.39}  
& \multicolumn{1}{c}{0.43}  
& \multicolumn{1}{c}{0.52}  
& \multicolumn{1}{c}{0.54}  
& \multicolumn{1}{c}{0.53} 
\\
\bottomrule
\end{tabular}
\caption{Results of prompt template in bangla and hindi of CREA-ICL method in SentNoB test of bloomz-3b.}
\label{tab:results of SentNoB test bloomz-3b}
\end{table*}

Expanding our scope beyond English, we explored the use of Bangla and its linguistically similar, high-resource counterpart Hindi as prompt templates \( P \), as detailed in Table \ref{tab:results of SentNoB test bloomz-3b}.


 We observed that using Hindi as a template language led to precision and recall improvements in some categories. However, it did not surpass the macro average F1 score achieved by the CREA-ICL method with English prompts. The Bangla template, while improving precision in some instances, suffered a drop in recall and overall accuracy, culminating in the least impressive macro average F1 score among the templates examined.

These results suggest that while the Bangla template may improve category-specific performance, it compromises the model's ability to generalize across the spectrum of categories in the SentNoB test. Similarly, the Hindi template's category-specific improvements in precision and recall do not translate into an increased macro-average F1 score over the CREA-ICL method with English prompts.

In conclusion, the comprehensive F1 score analysis underscores the superiority of the CREA-ICL method coupled with English prompts in terms of overall effectiveness. However, the impact of prompt template language on the performance of specific categories is significant, as evidenced by the Hindi and Bangla templates. This highlights the importance of striking a strategic balance between improving category-focused performance and maintaining overall effectiveness when selecting prompt templates for cross-language retrieval enhancement tasks.

\subsection{Impact of Hindi Retrieval Corpus}

Comparing the results in Table \ref{tab:task2 test hindi dataset} with the previous experiments, it is clear that neither Bloomz-3b nor mBERT show improvements over the CREA-ICL method using English Retrieval Corpus. This implies that while the use of different retrieval datasets has the potential to improve results within specific sentiment classifications, the selection of retrieval content must be carefully considered to optimize collective performance across different categories in cross-lingual sentiment analysis efforts.

A consistent challenge is the ``Neutral'' category using the Bloomz-3b model, which suffers from inadequate recall and F1 scores regardless of the retrieval corpus used. This pattern suggests that additional model refinements and strategy adjustments are needed to increase the accuracy of neutral sentiment retrieval.

\begin{table}[]
\centering
\begin{tabular}{ccccccc}
\toprule
             & \multicolumn{3}{c}{k=1}       & \multicolumn{3}{c}{k=3}       \\ \midrule
bloomz-3b    & precision & recall & f1-score & precision & recall & f1-score \\ \midrule
Negative     & 0.58      & 0.84   & 0.69     & 0.59      & 0.88   & 0.70     \\
Neutral      & 0.09      & 0.00   & 0.00     & 0.08      & 0.00   & 0.00     \\
Positive     & 0.55      & 0.49   & 0.52     & 0.58      & 0.47   & 0.52     \\
accuracy     &           &        & 0.57     &           &        & 0.58     \\
macro avg    & 0.41      & 0.44   & 0.40     & 0.42      & 0.45   & 0.41     \\
weighted avg & 0.48      & 0.57   & 0.51     & 0.49      & 0.58   & 0.51     \\ \toprule
mbert        & precision & recall & f1-score & precision & recall & f1-score \\ \midrule
Negative     & 0.48      & 0.24   & 0.32     & 0.48      & 0.33   & 0.39     \\
Neutral      & 0.21      & 0.34   & 0.26     & 0.21      & 0.28   & 0.24     \\
Positive     & 0.27      & 0.37   & 0.31     & 0.25      & 0.33   & 0.28     \\
accuracy     &           &        & 0.30     &           &        & 0.32     \\
macro avg    & 0.32      & 0.32   & 0.30     & 0.31      & 0.31   & 0.31     \\
weighted avg & 0.36      & 0.30   & 0.30     & 0.36      & 0.32   & 0.33     \\ \bottomrule
\end{tabular}

\caption{Results in SentNoB test of BLOOMZ-3b and mBERT with hindi retrieval corpus.}
\label{tab:task2 test hindi dataset}
\end{table}
\section{Conclusion}
In this study, we presented an innovative methodology CREA-ICL to harness the power of Large Language Models for low-resource languages, with particular emphasis on Bengali. By integrating cross-lingual retrieval-augmented in-context learning, we sought to enhance the capabilities of MPLMs, notably BLOOM and BLOOMZ. Our methods were rigorously evaluated across two classification tasks and two generation tasks.

The empirical outcomes underscore the success of our strategy, as evidenced by the notable F1-scores in classification tasks.  A deeper dive into our results reveals the pivotal role the CREA-ICL mechanism plays in strengthening the model's effectiveness.

Our research paves the way for ensuing investigations into cross-lingual retrieval and in-context learning's potential in the realm of low-resource languages. For future work, there's an exciting possibility to adapt and extend this framework to other marginalized languages and to address more complicated NLP challenges, including question-answering and machine translation.



{
\small
\bibliography{anthology,custom}
\bibliographystyle{unsrtnat}
}

\end{document}